\documentclass{article}

\usepackage{PRIMEarxiv}

\usepackage[utf8]{inputenc} 
\usepackage[T1]{fontenc}    
\usepackage{hyperref}       
\usepackage{url}            
\usepackage{booktabs}       
\usepackage{amsfonts}       
\usepackage{nicefrac}       
\usepackage{microtype}      
\usepackage{lipsum}
\usepackage{fancyhdr}       
\usepackage{graphicx}       
\graphicspath{{media/}}     
\usepackage{amsmath}
\title{Optimal consumption-investment choices under wealth-driven risk aversion

}

\author{
  Ruoxin Xiao\\
  Department of Mathematics
College of Sciences \\
  Shanghai University \\
  Shanghai, China\\
  \texttt{xiaoruoxin@shu.edu.cn} \\
}

\begin{document}
\maketitle

\begin{abstract}
 CRRA utility where the risk aversion coefficient is a constant is commonly seen in various economics models. But wealth-driven risk aversion rarely shows up in investor's investment problems. This paper mainly focus on numerical solutions to the optimal consumption-investment choices under wealth-driven aversion done by neural network. A jump-diffusion model is used to simulate the artificial data that is needed for the neural network training. The WDRA Model is set up for describing the investment problem and there are two parameters that require to be optimized, which are the investment rate of the wealth on the risky assets and the consumption during the investment time horizon. Under this model, neural network LSTM with one objective function is implemented and shows promising results. 
\end{abstract}


\section{Introduction}
The theory of risk aversion is developed to cope with the measurement of uncertainty in economics. In canonical theories, risk aversion is usually modeled by expected utility. The concept was first tied to diminishing marginal utility for wealth. In applications, economists derived specific functional forms to measure risk aversion. Two common measures are the coefficient of absolute risk aversion and the coefficient of relative risk aversion, both defined by Pratt (1964) and Arrow (1965). The constant relative risk aversion (CRRA) utility functon is one of the most widely used forms, in which risk aversion is modeled by a single constant parameter, $\rho$. 

Though the model performs well due to its simplicity, it still puts serious constraints on individual preferences. Much research have been done on modeling risk aversion in application. Risk aversion is studied in empirical analysis to be affected by exogenous factors, mainly demographic, measuring the heterogeneity of investors. In Palacios-Huerta and Santos (2004)\cite{PALACIOSHUERTA2004601}, they modeled the degree of risk aversion endogenous to market arrangements. To take a step further, this paper casts light on endogenous risk aversion model within the consumption-investment strategy problem.

Unlike the CRRA model where the risk aversion coefficient is constant, the innovation we present in this paper of consumption-investment optimization strategy is to take the influence of temporary wealth on risk aversion into consideration. Risk aversion is no longer fixed throughout the investment period, but a function that varies with the changes in temporary wealth, the result of optimal consumption-investment strategy of each step. The ultimate goal is to optimize the expected utility on consumption and terminal wealth through the risk-aversion-changing process.

The study of optimal consumption-investment problem can be traced back to Merton in 1969 (Merton, 1969, 1971). Merton develops an explicit optimal investment strategy by using stochastic optimal theory. Currently, deep learning skills have already been used in numbers of areas, including portfolio selection. Deep learning skills first used to solve optimal investment problem is done by Chen and Ge (2001).\cite{CHEN2021936}

\section{Jump-Diffusion Model and Wealth-driven Risk Aversion Model}
\label{sec:headings}

\subsection{Set up of the JD Model}
We begin with a financial market which operates continuously with a probability space 
($\Omega$,$\mathcal{F}$,$\mathcal{P}$) and a time horizon $0<t<T$. $\mathcal{F}_{t}$ is defined as a filtration reflecting all the available information at time t. We consider an investment universe consisting of one risk-free asset $P_{0}$ and one risky assets denoted by $S$. The price of assets $S$ is governed by the following stochastic differential equation\cite{Matsuda2004IntroductionTM}
\begin{equation}
dS_{t}=S_{t^{-}}(\mu dt+\sigma dB_{t}+dJ_{t})
\end{equation}
\begin{equation}
J_{t}=\sum _{i=1}^{N_{t}}(e^{U_{i}}-1)
\end{equation}
where $B_{t}$ is a standard Brownian Motion, $N_{t}$ is a Poisson process with rate $\lambda$ and $U_{i}$ is a sequence of independent identically distributed (i.i.d) random variables such that has an asymmetric double exponential distribution, which refers to Kou(2002)\cite{Kou2002AJM} with the density
\begin{equation}
    f_{Y}\left( y\right) = p\eta_{2}e^{-\eta _{1}\left( y-a\right) }1_ {\{y\geq  a\}} +q\eta_{2}e^{\eta _{2}\left( y-a\right) }1_ {\{y\geq  a\}}
\end{equation}
\begin{eqnarray*}
\eta_{2}>1,\eta_{2} >0
\end{eqnarray*}
where $p,q>0$, and $p+q=1$, representing the probabilities of upwards and downwards jumps compared with $y=\alpha$.

Solving the SDE(1) above by It$\hat{o}$'s Lemma gives the dynamics of an asset price
\begin{equation}
    S_t=S_0 exp((\mu-\frac{\sigma^2}{2})t+\sigma W_t+\sum_{i=1}^{N_t}U_i)
\end{equation}
\subsection{Set up of the WDRA Model}

Consider a financial market which consists of a riskless asset $P_{0}$ and a risky asset $S$. An investor enters the market at time 0 with initial wealth $w_{0}$. The price of the risk-free asset is governed by the following ordinary differential equation
\begin{equation}
dP_{0}=rP_{0}dt
\end{equation}
where $r$ is assumed to be constant which represents the risk-free rate. The price of risky assets $S$ is governed by the following stochastic differential equation
\begin{equation}
dS_{t}=S_{t^{-}}(\mu dt+\sigma dB_{t}+dJ_{t})
\end{equation}

We develop the binary utility function related to the state of wealth based on the Constant Relatively Risk Aversion (CRRA) utility function
\begin{equation}
u(x,y)=\frac{x^{1-\rho(y)}-1}{1-\rho(y)},\rho(y)\neq 1
\end{equation}
where $u(x,y)=log(x)$ if $\rho(y)=1$ and $\rho$ indicates the level of relative risk aversion. This wealth-driven risk aversion lets the coefficient of risk aversion varies with the wealth.

We adopt the function form of wealth-driven risk aversion developed by Chu, Nie and Zhang(2014)\cite{2013Wealth} through their empirical analysis. 
\begin{equation}
\rho(W)=b_{0}+b_{1}\cdot W+b_{2}\cdot W^{3}
\end{equation}
where $W$ is the ratio of individual wealth to the average wealth level and we set $b_{1}=0.13$, $b_{2}=-0.45$, being consistent with the empirical specification in Chu (2014). We chose $b_{0}$ such that the average risk aversion coefficient is 3.
The relationship between risk aversion and wealth is hump-shaped. Risk aversion first increases with wealth and then decreases with it, which suggests that the poorest group of people and the richest group of people are more of risk takers than the those people in the middle.

The investor is endowed with wealth $w_{0}$ at the beginning of the time horizon. And its objective is to maximize the the expected utility by making consumption and investment choices during the time horizon. The expected utility during the investment time horizon and the wealth process are given by the following formula
\begin{equation}
\mathop{sup}\limits_{\theta,c}E[\zeta\int_{0}^{T}e^{-\eta t}u(c_{t},w_{t})dt+(1-\zeta)e^{-\eta t}u(w_{T},w_{T})]
\end{equation}
\begin{equation}
dw_{t}=\theta_{t}w_{t}S^{-1}_{t}dS_{t}+(1-\theta_{t})w_{t}rdt-c_{t}dt
\end{equation}
where $\theta_{t}$ represents the investment rate of the wealth on the risky asset $S$ at time $t$, $(1-\theta_{t})$ represents the investment rate of the wealth on the riskless asset $P_{0}$ at time $t$, $c_{t}$ represents the consumption at time $t$, $w_{T}$ represents the terminal wealth, $\eta$ represents the subjective discount rate, and $\delta$ represents the relative importance of the intermediate consumption and the terminal wealth. $E$ is the expectation of the whole stochastic process. The first term and the second term in (9) is used to measure the utility in terms of consumption and terminal wealth under the wealth-driven risk aversion.
\subsection{Estimation of JD Model}

By(1), the log return over a time interval $\Delta t$ is:
\begin{equation}
\ln \left(\frac{S_{t+\Delta t}}{S_{t}}\right)=\left(\mu-\frac{\sigma^{2}}{2}\right) \Delta t+\sigma\left(W_{t+\Delta t}-W_{t}\right)+\sum_{i=N_{t}+1}^{N_{t+\Delta t}} U_{i}
\end{equation}
where we set the time interval is small($\Delta t$=one day= 1/247 year) to approximate 
the log return as
\begin{equation}
\ln \left(\frac{S_{t}+\Delta t}{S_{t}}\right) \approx\left(\mu-\frac{\sigma^{2}}{2}\right) \Delta t+\sigma \sqrt{\Delta t} Z+B Y
\end{equation}
where $Z$ is a standard normal random variables, $B$ is a Bernoulli random variable with $P(B=1)=\lambda \Delta t$ and $P(B=0)=1-\lambda \Delta t$. 

The density of (12) is given by the following formula
\begin{equation}
\begin{aligned}
g(x) &=\frac{1-\lambda \Delta t}{\sigma \sqrt{\Delta t}} \phi\left(\frac{t-\left(\mu-\frac{\sigma^{2}}{2}\right) \Delta t}{\sigma \sqrt{\Delta t}}\right) \\
&+\lambda \Delta t p \eta_{1} e^{\frac{\eta_{1}^{2} \sigma^{2} \Delta t}{2}} e^{-\left(t-\alpha-\left(\mu-\frac{\sigma^{2}}{2}\right) \Delta t\right) \eta_{1}} \Phi\left(\frac{t-\alpha-\left(\mu-\frac{\sigma^{2}}{2}\right) \Delta t-\eta_{1} \sigma^{2} \Delta t}{\sigma \sqrt{\Delta t}}\right) \\
&+\lambda \Delta t q \eta_{2} e^{\frac{\eta_{2}^{2} \sigma^{2} \Delta t}{2}} e^{\left(t-\alpha-\left(\mu-\frac{\sigma^{2}}{2}\right) \Delta t\right) \eta_{2}} \Phi\left(-\frac{t-\alpha-\left(\mu-\frac{\sigma^{2}}{2}\right) \Delta t+\eta_{2} \sigma^{2} \Delta t}{\sigma \sqrt{\Delta t}}\right)
\end{aligned}
\end{equation}

Firstly, to estimate an appropriate range of $\lambda$, we count the number of data points which are outside the $3\sigma$(Gaussian estimated sigma) interval. Then Maximum Likelihood Estimation is used to estimate seven unknown parameters$(\mu,\sigma,\lambda,p,\eta_{1},\eta_{2},\alpha)$ in the model by empirical data. The maximum likelihood function is given as follow
\begin{equation}
\max _{para}L\left( x_{1},x_{2},\ldots \cdot x_{n};para\right) =\max _{para }\sum ^{n}_{i=0}\ln g\left(  x_{i}| para \right) 
\end{equation}
where $\{x_{i}\}$ is empirical data processed from the data of a stock within one year and $para=(\mu,\sigma,\lambda,p,\eta_{1},\eta_{2},\alpha)$. During this process, Adam Optimizer is used to optimize the maximum likelihood function. As a result, the estimated parameters $(\mu,\sigma,\lambda,p,\eta_{1},\eta_{2},\alpha)$ is $($-0.2438$,0.2579,2,0.0062,1.0879,0.2435,0.2)$. The density plot of g is shown in Fig.1 along with the Gaussian kernel density estimation.
\clearpage
\begin{figure}[h]
    \centering
    \includegraphics[scale=0.8]{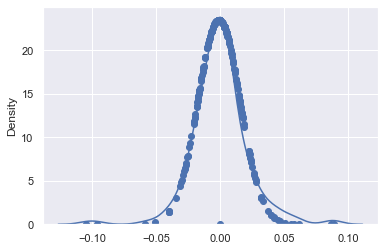}
    \caption{The fitted density plot of g is compared with the Guassian kernel density estimation. The blue points stands for the model and the blue line is used for the Guassian kernel density estimation by empirical data}
    \label{fig:my_label}
\end{figure}

\subsection{Simulation of JD model}

For the purpose of training neural networks, a set of data for the JD model is produced by simulation based on the estimated parameters in the previous section. The initial price is set at $S_{0}=100$. We simulated the prices of a hypothetical stock in $T=247$ days (one year) under the model for 100 times. The (12) could be divided into two part: the diffusion part and the jump part.
\begin{itemize}
\item The diffusion part: The diffusion part follows a normal distribution with mean $(\mu -\dfrac{\sigma^{2}}{2}) \Delta t$ and standard deviation $\sigma\sqrt{\Delta t}$.

\item The jump part:The jump part simulates the inter-jump time and the jump size. The inter-jump time follows the exponential distribution $Exp(\lambda)$, which indicates that the time at which the jumps occur follows a Poisson process with parameter $\lambda$. Accordingly, we iteratively draw samples from Poisson distribution till the sum of the samples exceeds 1 (since $T=1$ year). The sampled inter-jump times is stacked together to get the jump times and floored for specific dates. For the jump size, it follows a double-exponential distribution. The samples are draw from a double-exponential distribution for each time that the jump occurs and the number of the samples at each time is the inter-jump time.
\end{itemize}

The simulation paths is shown in Figure 2. From the figure, it is clear that there are both small and steady fluctuations and occasional steep changes in the paths.
\begin{figure}
    \centering
    \begin{tabular}{c}
    \includegraphics[scale=0.8]{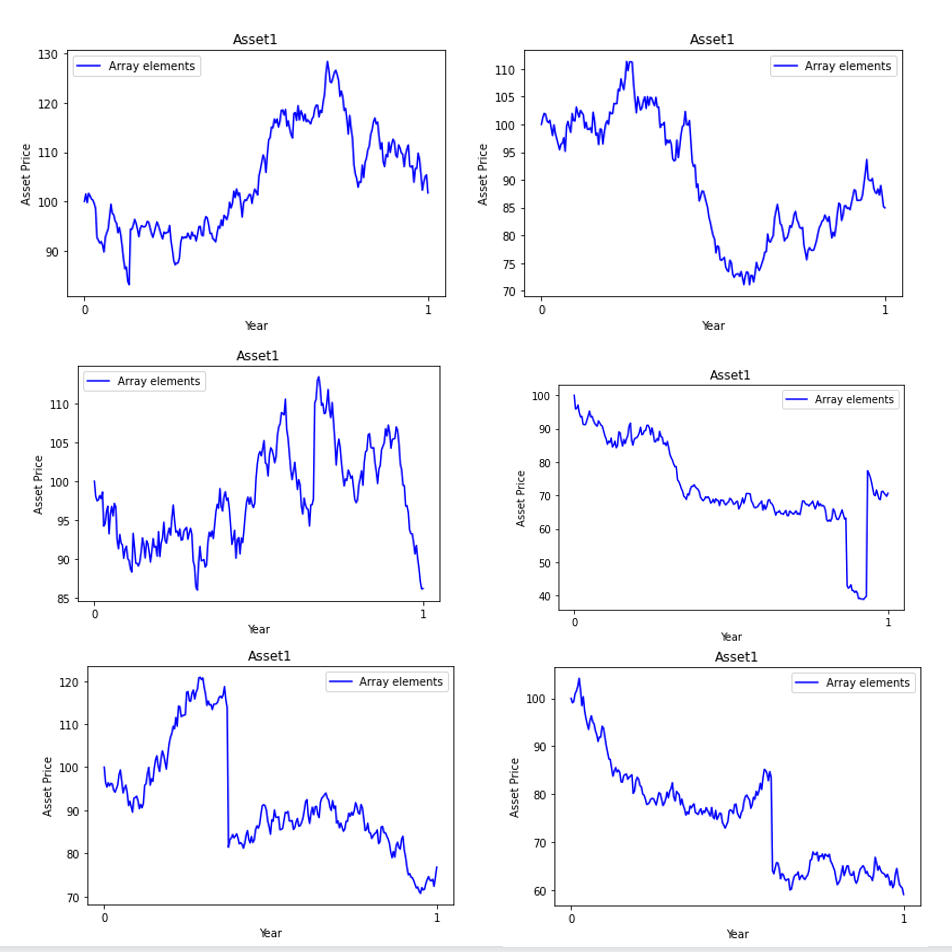}
    \end{tabular}
    \caption{Simulated paths for the JD Model}
    \label{fig:my_label}
\end{figure}
\clearpage

\subsection{LSTM}
Recurrent Neural Network (RNN) is a Neural Network used to process sequential data. For example, the meaning of a word may have different meanings depending on the content mentioned above, which could be solved by RNN. Long short-term memory (LSTM) is a special RNN, which is mainly used to solve the problem of gradient disappearance and gradient explosion in long sequence training. To put it simply, LSTM can perform better in longer sequences than ordinary RNN.\cite{kipf2018learned}

The core idea behind LSTM is the cell state, the horizontal line running across the top of the diagram.

The cell state is like a conveyor belt. It runs directly along the chain, with a few minor linear interactions. It's easy for information to flow through it unchanged.
LSTM has the ability to remove or add information to the cell state, carefully regulated by structures called gates.

Gates are a way of selectively letting information through. They consist of a sigmoid network layer and a pointwise multiplication operation.
The sigmoid layer outputs numbers between 0 and 1 describing how much each component should be allowed through. A value of 0 means let nothing through and a value of 1 means let everything through.

An LSTM has three such gates to protect and control the cell state, which are forget gate$\mathcal{F}_t$,input gate$\mathcal{I}_t$,output gate $\mathcal{O}_t$. The forget gate $\mathcal{F}_{t}$ decides which information should be removed from the memory state.The input gate $\mathcal{I}_t$ processed the new information and the output gate $\mathcal{O}_t$ evaluate the memory state and decide the outputtings.The gates control the information flow, y adding or moving information in the memory state $h_t$. The explicit operation of the gates follows\cite{fecamp2020deep}:
\begin{equation}
\begin{aligned}
\mathcal{F}_{t}=\sigma \left( Q_{F}S_{t}+R_{F}C_{t-1}+b_{F}\right) \\
\mathcal{I}_{t}=\sigma \left( Q_{I}S_{t}+R_I C_{t-1}+b_{I}\right) \\
\mathcal{O}_{t}=\sigma \left( Q_{O}S_t+R_{O}C_{t-1}+b_{O}\right) \\
h_t =\mathcal{F}_{t}\odot h_{t-1}+\mathcal{I}_{t}\odot \tanh \left(Q_{h}S_t+R_{h}C_{t-1}+b_{h}\right) ,h_0=0\\
C_t=\mathcal{O}_t\odot\tanh \left( h_t\right) ,C_0=0
\end{aligned}
\end{equation}
where $\odot$ is the Hadamard product, $\sigma$ is the sigmoid activation function ($\sigma \left( x\right) =\frac{1}{1-e^{-x}}$).
$Q_{l}$,$R_{l}$ are the affine transformations represented by
\begin{equation}
Q_{t}=W_{t}x+\beta_{t}
\end{equation}
$W_{t}x$ is the weight matrix and $\beta_{t}$ is the bias matrix.$Q_{\star}\in \mathbb{R}^{h\times d}$,$R_{\star}^{h\times h}$,$b_{\star}\in \mathbb{R}^h$
,h representing cell state size. Throughout the procedure, the weight matrices and bias vector ($Q_{\star}$,$R_{\star}$,$b_{\star}$) keep unchanged and are shared with all gates at each time step.
\begin{figure}[h]
    \centering
    \includegraphics[scale=0.3]{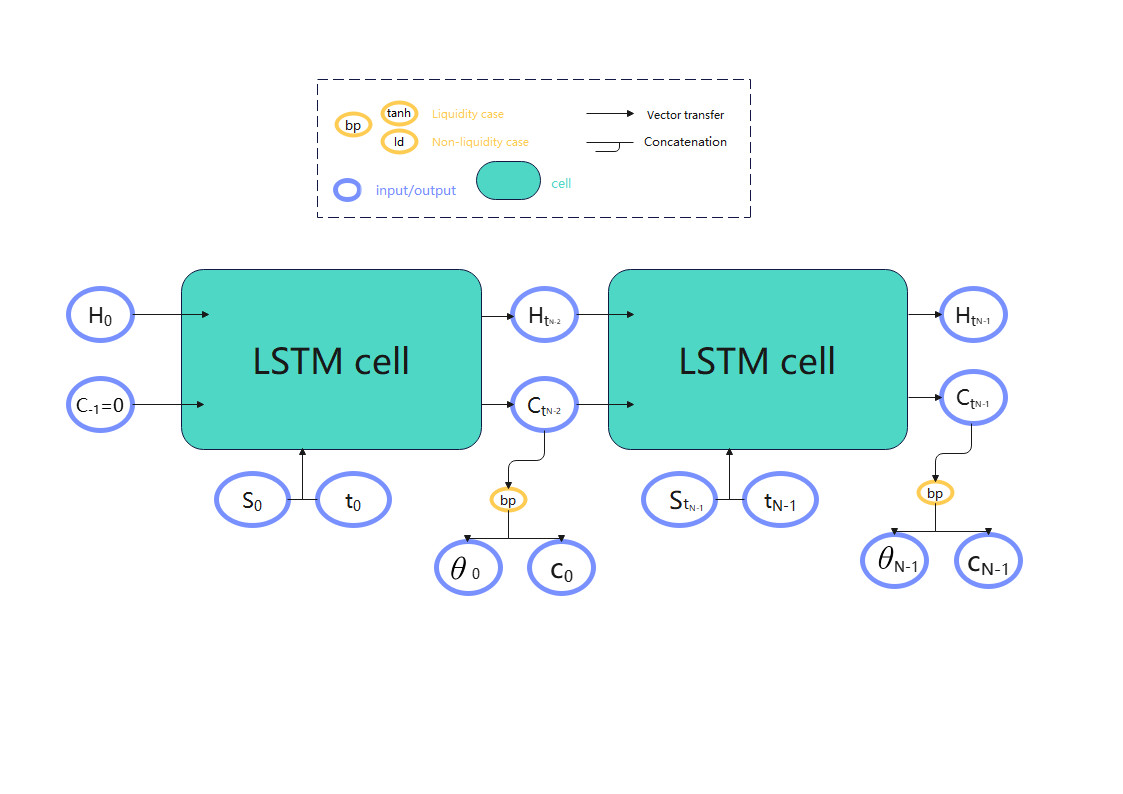}
    \caption{LSTM cell structure}
    \label{fig:my_label}
\end{figure}
\clearpage
\subsection{Adaptive Moment Estimation}
Adaptive Moment Estimation(Adam) is a first-order optimization algorithm that can replace the traditional stochastic gradient descent process. It can update the weight of neural network iteratively based on training data. Adam algorithm is different from traditional stochastic gradient descent. Stochastic gradient descent keeps a single learning rate (alpha) updating all weights, and the learning rate does not change during training. Adam designs independent adaptive learning rates for different parameters by calculating the first and second moment estimates of the gradient. Adam is a very popular algorithm in the field of deep learning because it can achieve excellent results very quickly. The empirical results show that Adam algorithm has excellent performance in practice and has great advantages over other kinds of random optimization algorithms.

\begin{center}

\begin{table}

\caption{Algorithm: Adaptive Moment Estimation}
\label{table1}

     1:\ $\alpha$:learning size\\
     2:\ $\beta_1,\beta_2 \in [0,1]$, Exponential dacay rates for the estimates\\
     3:\ $N_{epoch}$ number of epochs\\
     4:\ $N_{batch}$ the number of simulations at each gradient decent iteration(batch size)\\
     5:\ $\theta_0$ randomly chosen\\
     6:\ $m_0 \leftarrow 0$\\
     7:\ $v_0 \leftarrow 0$\\
     8:\ $t \leftarrow 0$\\
     9:\ for $t=0...N_{epoch}$\\
     10:$\qquad S_i \leftarrow N_{epoch}$ represents simulations of $S_i,i=t_0,...t_{N-1},T$\\
     11:$\qquad t \leftarrow t+1$\\
     12:$\qquad g_t=\nabla_{\theta}L(\mathbb{NN}^{\theta_{t-1}(S_i)-g(S_T)})$(get gradient w.r.t objective function)\\
     13:$\qquad m_t \leftarrow m_{t-1}+(1-\beta_1)g_t$(updated first moment estimate)\\
     14:$\qquad v_t \leftarrow \beta_2 v_{t-1}+(1-\beta_1){g_t}^2$(updated biased second raw moment estiamte)\\
     15:$\qquad \hat{m}_{t}=\frac{m_t}{1-\beta _{1}^{t}}$(computes bias-corrected first moment estimate)\\
     16:$\qquad \hat{v}_{t}=\frac{v_t}{1-\beta _{2}^{t}}$(computes bias-corrected second moment estimate ($\beta_i^{t}$ represents $\beta_i$ to the power of $t$, $i=1,2$)\\
     17:$\qquad \theta_t \leftarrow \theta_{t-1}-\alpha \hat{m}_{t}/\sqrt{\hat{v}_t}$(update parameters)
\end{table}
\end{center}
\subsection{Solving the Investment Problem with Neural Networks}
The neural network LSTM is fed with the data simulated in Section 2.3. For the investment problem described in Section 2, the wealth process with the initial wealth $w_{0}$ and the utility function under wealth-driven risk aversion is given by
\begin{equation}
dw_{t}=\theta_{t}w_{t}S^{-1}_{t}dS_{t}+(1-\theta_{t})w_{t}rdt-c_{t}(w_{t})dt
\end{equation}
\begin{equation}
u(x,y)=\frac{x^{1-\rho(y)}-1}{1-\rho(y)},\rho(y)\neq 1
\end{equation}
and we are presented with the following optimization problem
\begin{equation}
\theta^\ast,c^\ast=\mathop{sup}\limits_{\theta,c}E[\zeta\int_{0}^{T}e^{-\eta t}u(c_{t},w_{t})dt+(1-\zeta)e^{-\eta t}u(w_{T},w_{T})]
\end{equation}
where $\theta$ is a vector that contains investment rate of wealth on the risky asset at each time $t$ and $c$ is a vector that represents the consumption at each $t$ and is greater than zero.
The investment rate of wealth on the risky asset $\theta$ and the consumption $c$ are computed by the network. The output layer is connected to the two independent layers:
\begin{itemize}
    \item[*] Investment rate layer: a fully-connected layer with output size T; the result is transformed by sigmoid into a ratio vector, indicating the rate of wealth that invests in risky asset in each day. 
    \item[*] Consumption layer: a fully-connected layer with output size T; the result is transformed by ReLU into a vector with non-negative value, indicating at each $t$.
\end{itemize}
\subsection{Neural Network Hyper-parameters}
Here we list hyper-parameters in our neural network. 
\begin{itemize}
    \item The batch size is 10
    \item The network use the Adam optimizer, with initial learning rate $10^{-3}$
    \item Each network has one hidden layer with 50
    \item LSTM units
    \item Each network was trained for 1000 epochs
\end{itemize}
\section{Numerical Results for the Investment Problem}
\subsection{The results in WDRA Model}
The trend of expected utility trained by the network is shown below, which indicates that the expected utility improves as the iterative number increasing and finally nearly converges to the maximum.  The number of epochs is 1000. We also give the final wealth distribution, the final investment-consumption choices during the time horizon.\clearpage
\begin{figure}
    \centering
    \includegraphics[scale=0.8]{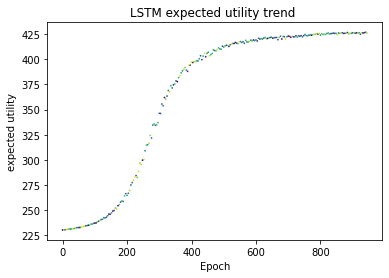}
    \caption{LSTM expected utility trend}
    \label{fig:my_label}
\end{figure}
\begin{figure}
    \centering
    \includegraphics[scale=0.8]{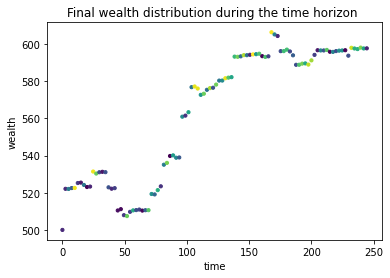}
    \caption{Final wealth distribution during the time horizon}
    \label{fig:my_label}
\end{figure}
\begin{figure}
    \centering
    \includegraphics[scale=0.8]{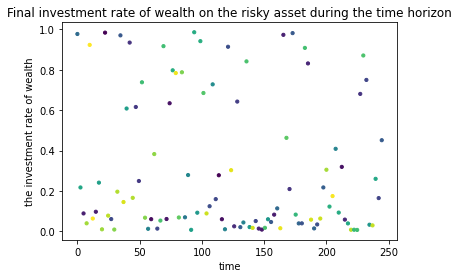}
    \caption{Final investment rate of wealth on the risky asset during the time horizon}
    \label{fig:my_label}
\end{figure}
\begin{figure}
    \centering
    \includegraphics[scale=0.8]{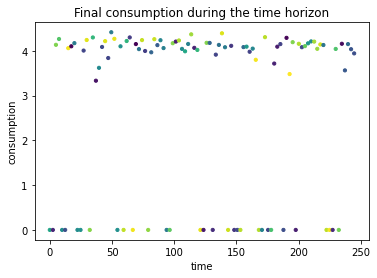}
    \caption{Final consumption during the time horizon}
    \label{fig:my_label}
\end{figure}
\clearpage
\subsection{Comparison of the WDRA Model and the CRRA Model}
CRRA Model means that the coefficient of risk aversion in the utility function is constant. The CRRA utility function is given as follow 
\begin{equation}
u(x)=\frac{x^{1-\rho}-1}{1-\rho},\rho\neq 1
\end{equation}
where $u(x)=log(x)$ if $\rho=1$ and $\rho$ is a constant and indicates the level of relative risk aversion.
In our paper, we mainly focus on the WDRA Model where the utility function is binary and related to the state of wealth
\begin{equation}
u(x,y)=\frac{x^{1-\rho(y)}-1}{1-\rho(y)},\rho(y)\neq 1
\end{equation}
where the coefficient of risk aversion varies with the wealth.
Here we compare the results of the expected utility, the final wealth distribution and  the final investment-consumption choices during the time horizon done by neural network LSTM.
\clearpage
\begin{figure}
    \centering
    \includegraphics[scale=0.8]{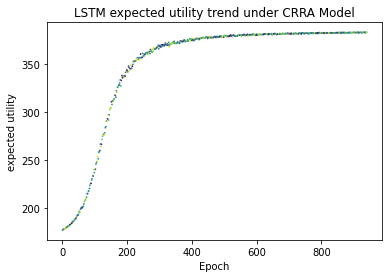}
    \caption{LSTM expected utility trend under CRRA Model}
    \label{fig:my_label}
\end{figure}
\begin{figure}
    \centering
    \includegraphics[scale=0.8]{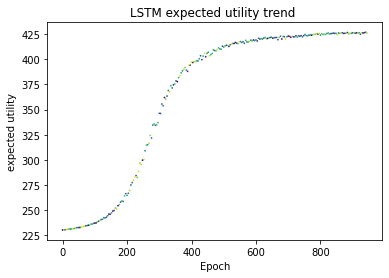}
    \caption{LSTM expected utility trend under WDRA Model}
    \label{fig:my_label}
\end{figure}
\begin{figure}
    \centering
    \includegraphics[scale=0.8]{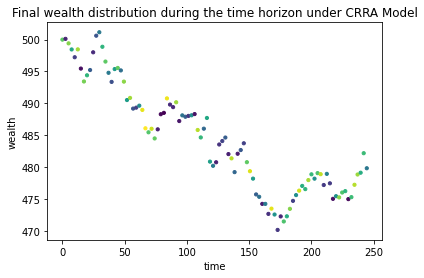}
    \caption{Final wealth distribution during the time horizon under CRRA Model}
    \label{fig:my_label}
\end{figure}
\begin{figure}
    \centering
    \includegraphics[scale=0.8]{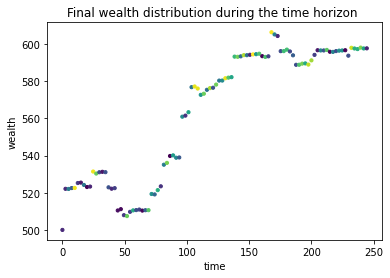}
    \caption{Final wealth distribution during the time horizon under WDRA Model}
    \label{fig:my_label}
\end{figure}
\begin{figure}
    \centering
    \includegraphics[scale=0.8]{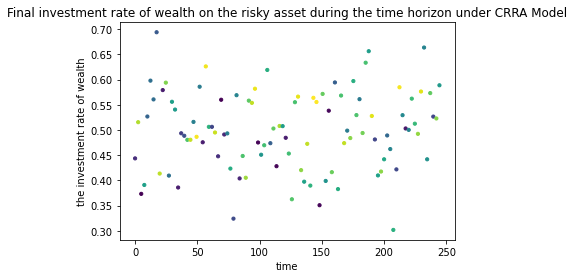}
    \caption{Final investment rate of wealth on the risky asset under CRRA Model}
    \label{fig:my_label}
\end{figure}
\begin{figure}
    \centering
    \includegraphics[scale=0.8]{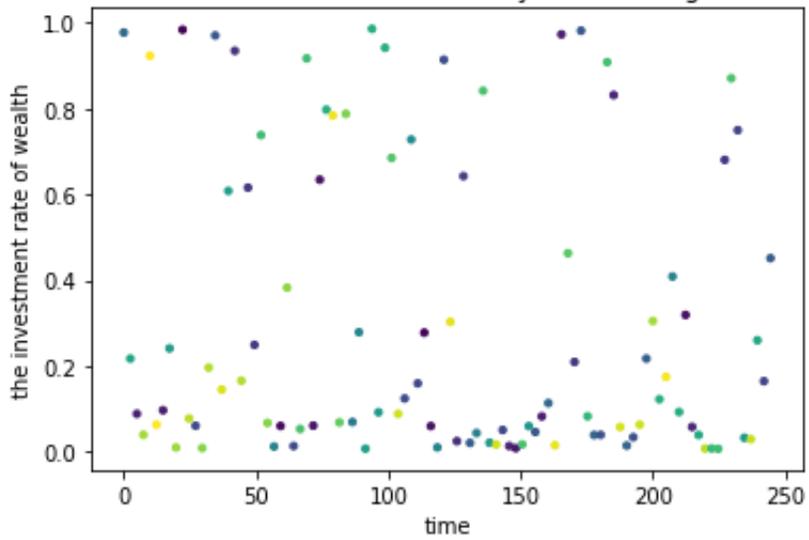}
    \caption{Final investment rate of wealth on the risky asset under WDRA Model}
    \label{fig:my_label}
\end{figure}
\begin{figure}
    \centering
    \includegraphics[scale=0.8]{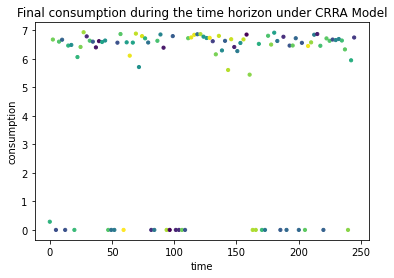}
    \caption{Final consumption during the time horizon under CRRA Model}
    \label{fig:my_label}
\end{figure}
\begin{figure}
    \centering
    \includegraphics[scale=0.8]{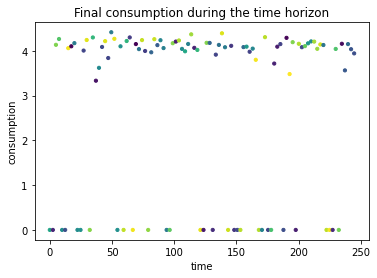}
    \caption{Final consumption during the time horizon under WDRA Model}
    \label{fig:my_label}
\end{figure}
\clearpage

Figure 8,9 shows that the speed of convergence in CRRA Model is greater than that in WDRA Model and the maximal expected wealth-driven utility is relatively larger than the the normal one.Figure 12,13 show that the rates of investment on risky asset is concentrated in the middle, close to 0.5 in CRRA Model while they fluctuates greatly and most of them are distributed at both extreme of the interval in WDRA Model. Figure 10,11,14,15 indicates that in the case the investor tends to consume under normal CRRA utility function is favourable to keep the wealth with wealth-driven utility function.
\section{conclusion}

In this paper, we extend the normal CRRA utility function to the wealth-driven one. To training the WDRA Model by neural network, a set of artificial stock data is simulated under the jump-diffusion model for 100 times. In addition, for the purpose of estimating the parameters in JD Model, we use the method of Maximum Likelihood Estimation and optimize the maximal likelihood function with Adam optimizer. Then LSTM network is used to solve the investment problem. It is fed with the artificial data obtained in the simulation part. Through the neural network, it is found that the expected utility is convergent to the maximum under both the WDRA Model and the CRRA Model. By comparing two models, we conclude some difference from the convergence and maximum of the expected utility, characteristics of investment behavior and the conditions of wealth and consumption.

We show that the use of deep learning neural network to solve the investment problem under wealth-driven risk aversion.The application of neural networks in other utility models and further interpretation could be improved in the future.
\clearpage

\bibliographystyle{ieeetr}

\bibliography{SHU}

\end{document}